\documentclass[preprint,12pt]{elsarticle}

\usepackage{amssymb}
\usepackage{graphicx}
\usepackage{amsthm}
\usepackage{amsmath}
\usepackage{natbib}
\usepackage{booktabs}

\journal{Arxiv}

\begin{document}

\begin{frontmatter}



\title{Hyper-parameter estimation method with particle swarm optimization}


\author{Yaru Li, Yulai Zhang, Xiaohan Wei}

\address{Department of Software Engineering, Zhejiang University of Science and Technology, Hangzhou, China, 310023, zhangyulai@zust.edu.cn
}

\begin{abstract}

\noindent Particle swarm optimization (PSO) methods cannot be directly used in the problem of hyper-parameters estimation since the mathematical formulation of the mapping from hyper-parameters to loss function or generalization accuracy is unclear. Bayesian optimization (BO) framework is capable of converting the optimization of hyper-parameters into the optimization of an acquisition function. The acquisition function is  non-convex and multi-peak. So the problem can be better solved by the PSO. The proposed method in this paper uses the particle swarm method to optimize the acquisition function in the BO framework to get better hyper-parameters. The performances of proposed method in both of the classification and regression models are evaluated and demonstrated. The results on several benchmark problems are improved.

\end{abstract}


\end{frontmatter}

\section{Introduction}

\noindent Particle swarm optimization (PSO) \cite{bai2010analysis} methods have been successfully used in the estimation of model parameters in the field of machine learning \cite{regulski2014estimation}\cite{schwaab2008nonlinear}. However, when it comes to the problem of hyper-parameters estimation \cite{bergstra2011algorithms}\cite{krajsek2006marginalized}, particle swarm methods, as well as many other optimization methods, cannot be directly used to deal with this problem. The difficulty lies in the fact that the mapping from hyper-parameters of  model to the loss function or generalization error is lack of explicit mathematical expressions and the computation complexity is very high.

Therefore, naive methods such as grid search \cite{stoica1999maximum} and random search \cite{bergstra2012random} are always used in the field of hyper-parameters estimation   in traditional engineering practices. These methods run large numbers of independent experiments under different hyper-parameter guesses and then pick the best hyper-parameters. Recently, bayesian optimization (BO) framework \cite{frazier2018tutorial}\cite{shahriari2015taking} is proposed in to deal with the hyper-parameters estimation problem in the machine learning community. BO tries to find the optimal value of a black-box function by constructing a posterior probability of the black box function's output when a finite number of the sample points are obtained from experiments. A surrogate model is used to construct the mapping from the hyper-parameters to the model accuracy. Then, it turns the optimization of the hyper-parameters into an optimization problem of the acquisition function \cite{frazier2018tutorial}\cite{shahriari2015taking}. The acquisition function describes the likelihood of the maximum or minimum points of the generalized accuracy or error of the model. The mathematical expression of the acquisition function may be high dimensional and has plenty of local minimum points.

The particle swarm method fits well with the above task. In this paper, we use the PSO method to solve the optimization problem of the acquisition function. In the existing works, the gradient based optimization methods such as L-BFGS-B \cite{liu1989limited}and TNC \cite{nash2000survey} are used in the BO framework. Calculating the first derivatives and the second derivatives of the acquisition functions are computationally expensive and the global optimal cannot be well guaranteed. If the PSO can return better results on the acquisition functions, the generalized accuracy of the machine learning model can be improved with high probability.

In the rest of this paper. The preliminaries of the PSO and BO are described in section 2; The section 3 introduces in details of the proposed method; The section 4 verifies the performance of the algorithm by experiments; Conclusions are offered in section 5.

\section{Preliminaries}\label{Preliminaries}

\subsection{Particle Swarm Optimization}

\noindent PSO is a method based on swarm intelligence, which was first proposed by Kenndy and Eberhart in 1995 \cite{bai2010analysis}\cite{hung2009hybridization}. Because of its simplicity in implementation, PSO algorithm is successfully used in machine learning, signal processing, adaptive control and so on \cite{wenjing2007parameter}.

As first step, a population of $m$ particles is initialized randomly, each particle is a potential solution to the problem that needs to be solved in the search space. In each iteration, the velocities and positions of each particle are updated using two values: one is the  best value $(p_b)$ of particle, and the other is the best value $(g_b)$ of population overall previous . Suppose there are $m$ particles in the $d$-dimensional search space, the velocity and position of the $i$-th particle at the time of $t$ are expressed as

\begin{equation}\nonumber
    v_i(t)=[v_{i1}(t),v_{i2}(t),\cdots,v_{id}(t)]^T
\end{equation}
\begin{equation}\nonumber
    x_i(t)=[x_{i1}(t),x_{i2}(t),\cdots,x_{id}(t)]^T
\end{equation}
the best value of particle and the overall previous best value of population at iteration $t$ are

\begin{equation}\nonumber
    p_{bi}(t)=[p_{i1}(t),p_{i2}(t),\cdots,p_{id}(t)]^T
\end{equation}
\begin{equation}\nonumber
    g_b(t)=[g_1(t),g_2(t),\cdots,g_d(t)]^T
\end{equation}
At iteration $t+1$, the position and velocity of the particle are updated as follows:

\begin{equation}\label{eiv}
     v_i(t+1)=\omega v_i(t)+c_1 r_1( p_{bi}(t)-x_i(t))+c_2 r_2(g_b(t)-x_i(t))
\end{equation}
\begin{equation}\label{eiv}
     x_i(t+1)=x_i(t)+v_i(t+1)
\end{equation}
where $\omega$ is the inertia weight coefficient, which can trade off the global search ability against local search ability;  $c_1$ and $c_2$ are the learning factors of the algorithm. If $c_1=0$, it is easy to fall into local optimization and can not jump out; if $c_2=0$, it will lead to slow convergence speed of PSO; $r_1$ and $r_2$ are random variables uniformly distributed in $[0, 1]$.

In each iteration of the PSO algorithm, only the optimal particle can transmit the information to other particles. The algorithm generally has two termination conditions: a maximum number of iterations or a sufficiently good fitness value. The process of PSO is as follows\\

\noindent {Algorithm $1$ Particle Swarm Optimization}

\noindent Step 1. Initialize a population of particles with random position and velocity on $d$-dimensions of the problem space;

\noindent Step 2. For each particle, evaluate the fitness function of each particle in $d$-dimensional;

\noindent Step 3. Compare particle’s fitness evaluation with particle’s $p_b$ . If current value is better than $p_b$, then let $p_b$ equal to the current value, and the $p_b$ location equal to the current location;

\noindent Step 4. Compare fitness evaluation with the population’s overall previous best. If current value is better than $g_b$,then reset $g_b$ to the current particle’s array index and value;

\noindent Step 5. Change the position and velocity of the particle according to equations $(1)(2)$;

\noindent Step 6. Go to Step $2$ until a criterion is met.

\subsection{Bayesian optimization}

BO was first proposed by Pelikan of the University of Illinois at Urbana-Champaign in 1998 \cite{snoek2012practical}. Under the condition that the finite sample points are known, BO finds the optimal value of the function by constructing a posterior probability of the output of the objective function $f$ \cite{frazier2018tutorial}\cite{shahriari2015taking}. Because the BO framework is very data efficient, it is particularly useful in situations where evaluations of $f$ are costly and one does not have access to derivatives with respect to $x$ and $f$ is non-convex and multi-peak. BO framework has two key ingredients. One is a probabilistic surrogate model, which consists of a prior distribution. The other is an acquisition function. BO is a sequential model-based approach since the postierior propability is constructed consequently by each data points \cite{brochu2010tutorial}.

Mathematically, denote $f(x)$ as objective function

\begin{equation}\label{py}
    x^\ast=argmaxf(x)
\end{equation}
where $x \in X$, $X \subseteq R^d$, $X$ is hyper-parameters space. The purpose of this article is to find the maximum value of the objective function. Suppose the existing data is $D_{1:t}={(x_i,y_i)}, i=1,2,\cdots,t$, $y_i$ is the generalization accuracy of the model under the hyper-parameter $x_i$. In the following, $D_{1:t}={(x_i,y_i)}, i=1,2,\cdots,t$ was simplified as $D$. We hope to estimate the maximum value of the objective function in a limited number of iterations. If $y$ is regarded as a random observation of the generalization accuracy, $y=f(x)+\varepsilon$, where the noise $\varepsilon$ satisfies $p(\varepsilon)=N(0,\sigma_\varepsilon^2)$, i.i.d.. The goal of hyper-parameter estimation is to find  $x^\ast$ in the $d$-dimensional hyper-parameters space.

One problem with this maximum expected accuracy framework is that the true sequential accuracy is typically computationally intractable. This has led to the introduction of many myopic heuristics known as acquisition functions, which is maximized as

\begin{equation}\label{py}
    x_{t+1}=argmax\quad\alpha_t(x;D)
\end{equation}
There are three commonly acquisition functions: probability of improvement (PI), expected improvement (EI) and upper confidence bounds (UCB). These acquisition functions trade off exploration against exploitation.

In recent years, BO has been widely used in machine learning model hyper-parameters estimation and model automatic selection \cite{mahendran2012adaptive}\cite{hennig2012entropy}\cite{garrido2020dealing}\cite{toscano2018bayesian}\cite{seeger2004gaussian}, which promotes the research of BO method for hyper-parameters estimation in many aspects.

\section{PSO-BO}\label{PSO-BO method}

The BO algorithm based on PSO is an iterative process. First of all, use Algorithm $1$ to optimize the acquisition function to obtain $x_{t+1}$; Then, evaluate the objective function value according to $ y_{t+1}=f(x_{t+1})+\varepsilon$; Finally, update $D$ with the new sample point $\{(x_{t+1},y_{t+1})\}$, and update the posterior distribution of the probabilistic surrogate model for the next iteration.

\subsection{Algorithm Description}

The effectiveness of BO depends on the acquisition function $\alpha$. In general, $\alpha$ is non-convex and multi-peak, which needs to solve the non-convex optimization problems in the search space $X$. PSO algorithm is simple, with a few adjustment parameters and fast convergence speed. It is not necessary to calculate the derivatives of the objective function in the process of PSO. Therefore, PSO algorithm was chosen to optimize acquisition function to obtain new sample point in this paper.

The first choice we need to make is the surrogate model. Using a Gaussian process (GP) as the surrogate model is a popular choice, due to the potent function approximation properties and ability to quantify uncertainty of GP. A GP is a prior over functions which allows us to encode our prior beliefs about the properties of the function $f$, such as smoothness and periodicity \cite{bergstra2011algorithms}. GP is a nonparametric model \cite{srinivas2009gaussian} that is fully characterized by its prior mean function and its positive-definite kernel, or covariance function. Formally, each finite subset of GP model obeys multivariate normal distribution. Assuming that the output expectation of the model is $0$, the joint distribution of the original observation data $D$ and the new sample point $(x_{t+1},y_{t+1})$ can be expressed as follows

\begin{center}
$[y_{1:t+1}] \sim N\left(0, \left[ {\begin{array}{*{20}{c}}
  { K+\sigma_{\varepsilon}^2I} & \textbf k  \\
   {\textbf k^T} & {k(x_{t+1},x_{t+1})}  \\
 \end{array} } \right]\right)$
\end{center}
where $k:x*x\rightarrow\mathbb{R}$ is the covariance function, $ \textbf{k}=[k(x_1,x_{t+1}),\cdots,k(x_t,x_{t+1})]^T$
Gram matrix
\begin{center}
$K=\left[ {\begin{array}{*{20}{c}}
   {k(x_{1},x_{1})} & \cdots &{k(x_{1},x_{t})}   \\
         \vdots  & \ddots &  \vdots   \\
   {k(x_{t},x_{1})} & \cdots &{k(x_{t},x_{t})}  \\
 \end{array} } \right]$
\end{center}
$I$ is the identity matrix and $\sigma_\varepsilon^2$ is the noise variance. The prediction can be made by considering the original observation data as well as the new $x$. Since the posterior distribution of $y_{t+1}$ is

\begin{equation}\label{sample}
    p(y_{t+1}\mid y_{1:t},x_{1:t+1})=N(\mu_{t}(x_{t+1}),\sigma_{t}^2(x_{t+1}))
\end{equation}
The mathematical expectation and variance of $y_{t+1}$ are as follows

\begin{equation}
   \mu_{t}(x_{t+1}) = k^T(K+\sigma_{\varepsilon}^2I)^{-1}y_{1:t}
\end{equation}
\begin{equation}
   \sigma_{t+1}=k(x_{t+1},x_{t+1})-k^T(K+\sigma_{\varepsilon}^2I)^{-1}k
\end{equation}
The ability of GP to express the distribution of functions only depends on the covariance function. Matern-$52$ covariance function is one of them and as follows

\begin{equation}\label{sample}
    K_{M52}(x,x')=\theta_0\left(1+\sqrt{5r^2(x,x')}+\frac{5}{3}r^2(x,x')\right)exp\{-\sqrt{5r^2(x,x')}\}
\end{equation}

The second choice we need to make is acquisition function. Although our method is applicable to most acquisition functions, we choose to use UCB which is more popular in our experiment. GP-UCB proposed by Srinivas in 2009 \cite{srinivas2009gaussian}. The UCB strategy considers to increase the value of the confidence boundary on the surrogate model as much as possible, and its acquisition functions is as follows
\begin{equation}\label{sample}
    \alpha_{UCB}(x)=\mu(x)+\gamma\sigma(x)
\end{equation}
$\gamma$ is a parameter that controls the trade-off between exploration (visiting unexplored areas in $X$) and exploitation (refining our belief by querying close to previous samples). This parameter can be fixed to a constant value.

\subsection{Algorithm framework}

PSO-BO consists of the following  steps: (i) assume a surrogate model for the black box function $f$ , (ii) define an acquisition function $\alpha$ based on the surrogate model of $f$, and maximize $\alpha$ by the PSO to decide the next evaluation point , (iii) observe the objective function at the point specified by $\alpha$ maximization, and update the GP model using the observed data. PSO-BO algorithm repeat (ii) and (iii) above until it meets the stopping conditions. The algorithm framework is as follows\\

\noindent  {Algorithm $2$ PSO-BO}

\noindent Input: surrogate model for $f$, acquisition function $\alpha$

\noindent Output: hyper-parameter vector optimal $x^*$

\noindent Step 1. Initialize hyper-parameter vector $x_0$;

\noindent Step 2. For $t=1,2,...,T$ do:

\noindent Step 3. Using algorithm $1$ to maximize the acquisition function to get the next evaluation point: $x_{t+1}=argmax_{x\in X}\alpha(x|D)$;

\noindent Step 4. Evaluation objective function value $y_{t+1}=f(x_{t+1})+\varepsilon_{t+1}$;

\noindent Step 5. Update data:$D_{t+1}=D\cup(x_{t+1},y_{t+1})$, and update the surrogate model;

\noindent Step 6. End for.

\section{Experiments}\label{experiments}

PSO can guarantee the convergence of the algorithm when choosing $(\omega,c_1,c_2)$ within this stability region: $-1<\omega<1$, $0<c_1+c_2<4(1+\omega)$ \cite{jiang2007stochastic}\cite{zheng2003convergence}. $c_1$ and $c_2$ affect the expectation and variance of position. The smaller the variance is, the more concentrated the optimization results are, and the better the stability of the optimization system is. \cite{bai2010analysis} have studied the influence of values of $c_1$, $c_2$ on the expectation and variance of position for the purpose of reducing variance.

In order to demonstrate the performances of the proposed PSO-BO algorithm, two different data sets were analyzed on AdaBoost regressor, random forest (RF) classifier and XGBoost classifier \cite{chen2016xgboost}. Both of the datasets were randomly split into train/validation sets. The zero mean function and Matern-$52$ $(8)$ covariance function were adopted as the prior for the GP. In experiments we used the AdaBoost regressor, RF classifier and XGBoost classifier implementation available in Scikit-learn.

\subsection{Data sets and setups}\label{Data sets and setups}

Two datasets are used in this experiment: Boston housing dataset and Digits dataset \cite{hettich1998uci} , which are available in Scikit-learn. The Boston housing dataset contains $506$ examples and each example has $13$ dimensions for regression tasks. The Digits dataset contains $1797$ examples and each example has $64$ dimensions for classification tasks.

We first estimated four hyper-parameters of RF classifier model trained on the Boston housing dataset to set the value of $\omega$ . A table with the hyper-parameters to be optimized, their range and their type is displayed in Table $1$. Note that while the first parameter takes real values, the others take integer values. In the process of optimizing the hyper-parameters of RF classifier by PSO-BO, $\omega$ was set from $0.1$ to $0.9$, and all other parameters were set to the same to ensure a fair comparison. The experiments were repeated $10$ times. We use accuracy to measure performance on the classification task and the averaged results were shown in Fig.$1$. The vertical axis represents the accuracy on the validation set of the Boston housing dataset, and $5$-fold cross validation was used on the dataset. As Fig.1 is shown, when $\omega=0.8$, the value of accuracy is the highest, and as $\omega$ growing, however, the time required for the process of optimizing also increases significantly. Refer to the research of  \cite{bai2010analysis} on parameter setting of PSO algorithm, that PSO has the biggest rate of convergence when $\omega$ is between $0.8$ and $1.2$,  the parameter of PSO algorithm in PSO-BO was set to $c_1=1.85$, $c_2=2$, $\omega=0.8$.

\begin{center}
\begin{table}[htbp]\normalsize
 \caption{Types and range of hyper-parameters of RF } \centering

 \begin{tabular}{ccccccc}  \hline
\textbf {Name} & \textbf {Type}  & \textbf {Range} & \\
\hline
Max features & Real & (0.1, 0.999) & \\
Number of estimators & Integer & (10, 250) & \\
Minimum Number of Samples to Split & Integer & (2, 25) & \\
Max depth & Integer & (5, 15) & \\ \hline
 \end{tabular}\label{}
\end{table}
\end{center}

\begin{figure}[!htb]
\centering
  \includegraphics[width=245pt]{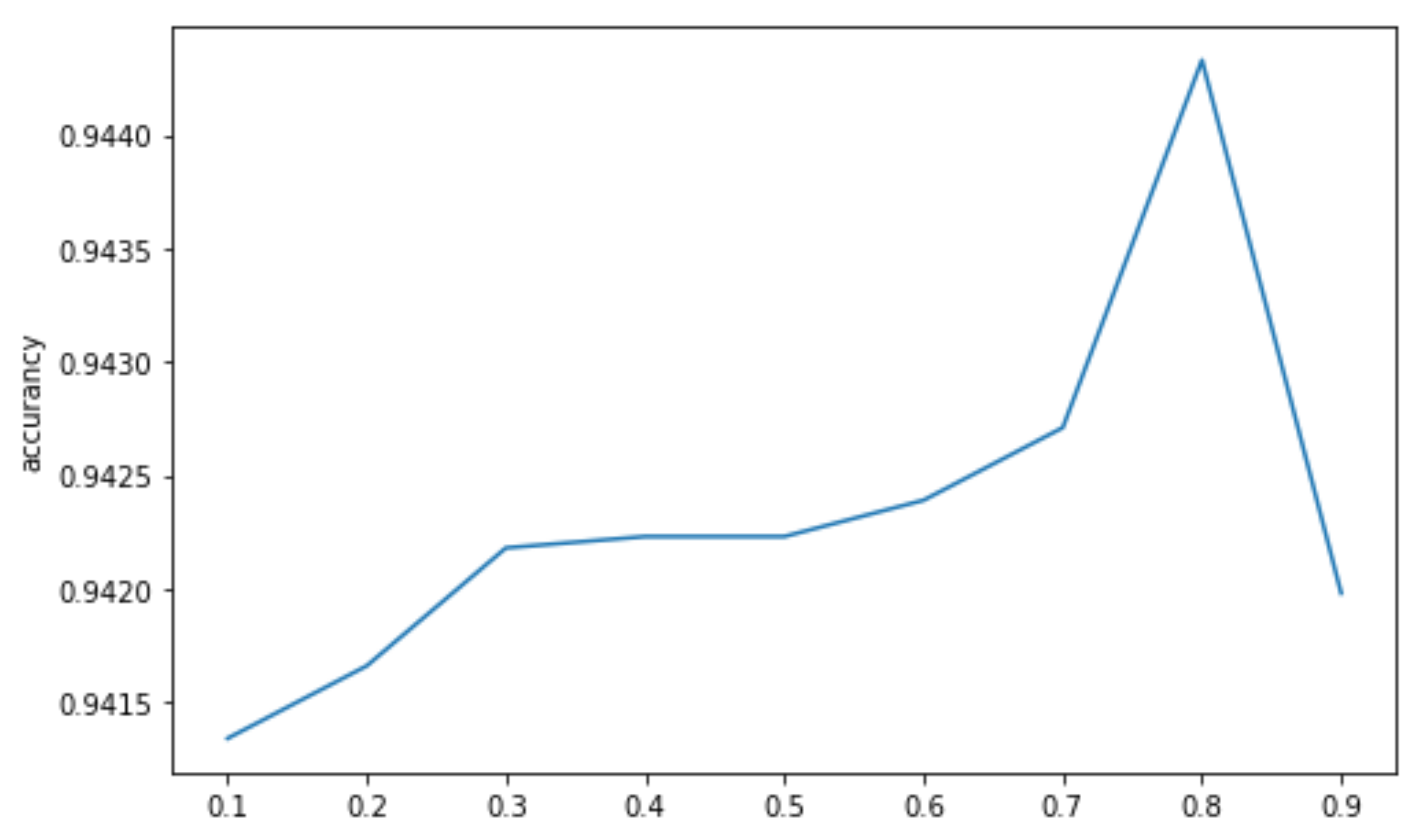}\\
  \caption{Experimental results of the developed PSO-BO approach with different values of $\omega$. Horizontal-axis: values of $\omega$ ; Vertical-axis: the $accuracy$ on the validation set of RF classifier model. }\label{f1}
\end{figure}

In the process of optimizing hyper-parameters of machine learning, there are generally bounds on hyper-parameters. However, among the methods used to optimize the acquisition function in the Bayesian optimization framework, L-BFGS-B, TNC, SLSQP, and trust-constr method apply to this restriction, in which L-BFGS-B and TNC are the most commonly used methods. Then, in this paper, L-BFGS-B and TNC are selected as the methods to optimize the acquisition function in the comparative experiment, and the corresponding Bayesian optimization framework is referred to as L-BFGS-B-BO and TNC-BO respectively.

To ensure a fair comparison, we implemented all methods in Python using the same packages. In all experiments, we used a zero-mean GP surrogate model with a Matern-$52$ kernel. We optimized the kernel and likelihood hyper-parameters by maximising the log marginal likelihood \cite{bai2010analysis}. All methods used UCB as the acquisition function. Each experiment was started with $5$ random initial observations.

\subsection{Hyper-parameter Optimization of AdaBoost}

In this section, we estimated two hyper-parameters $(d=2)$ of AdaBoost regressor using Boston housing dataset. And the two hyper-parameters were estimated by PSO-BO, L-BFGS-B-BO and TNC-BO respectively. The range and type of hyper-parameters be optimized is displayed in Table $2$. Note that while the first parameter takes real values, the other takes integer values. We use $R2$ to measure performance on the regression task. As shown in Table $3$, comparing the averaged results, maximum results and minimum results, PSO-BO outperforms the other two algorithms under comparison in terms of the $R2$ on the validation set of the Boston housing dataset, and $5$-fold cross validation was used on the dataset. In Fig.$2$, the vertical axis represents the $R2$, it can be seen that PSO-BO with $\omega=0.8$  performed better than the other settings.

\begin{center}
\begin{table}[htbp]\normalsize
 \caption{Types and range of hyper-parameters of AdaBoost } \centering

 \begin{tabular}{ccccccc}  \hline
\textbf {Name} & \textbf {Type}  & \textbf {Range} & \\
\hline
Learning rate & Real  & (0.1, 1) & \\
Number of estimators & Integer & (10, 250) & \\ \hline
 \end{tabular}\label{}
\end{table}
\end{center}

\begin{figure}[!htb]
\centering
  \includegraphics[width=245pt]{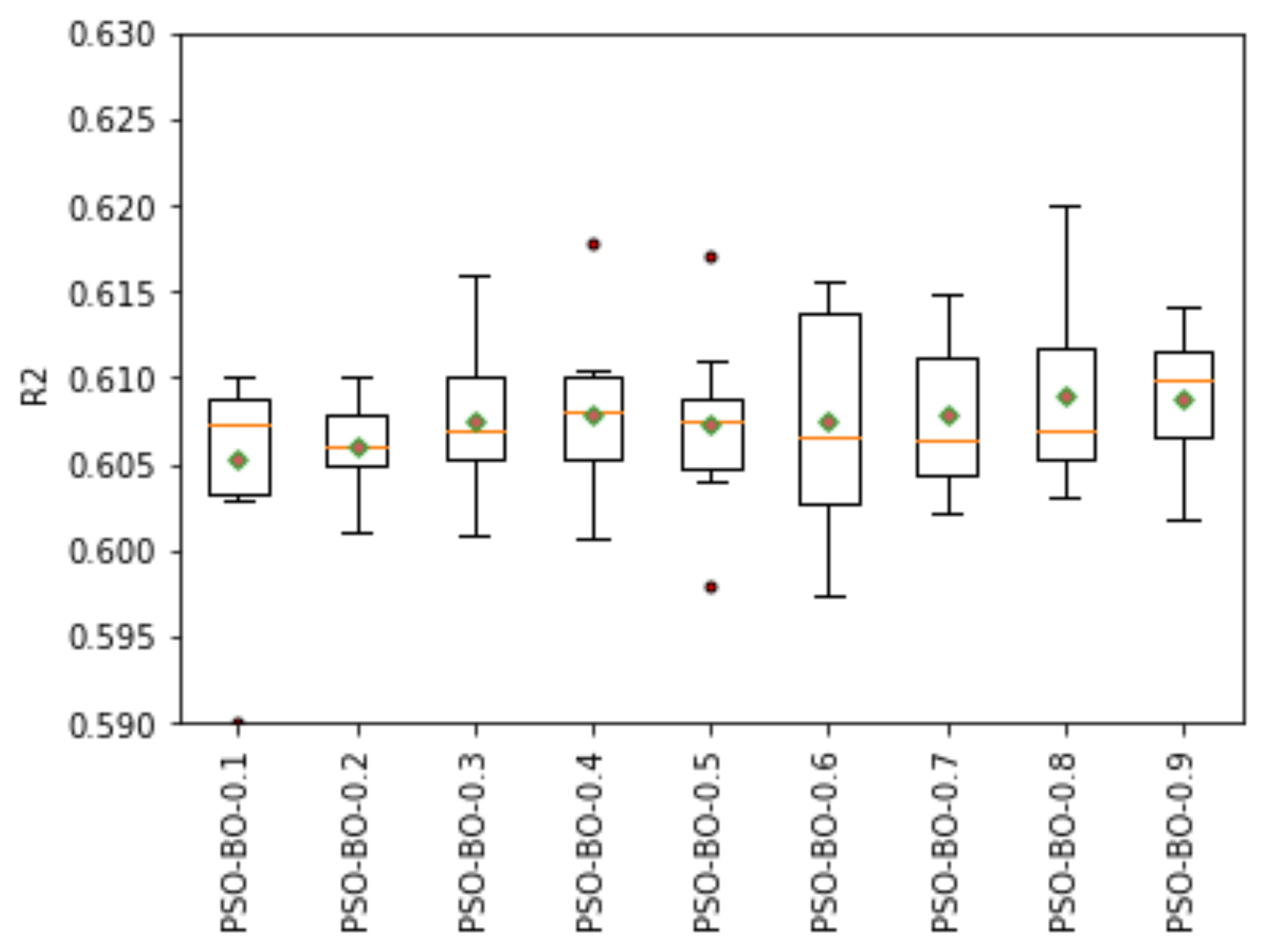}\\
  \caption{Experimental results of the developed PSO-BO approach with different values of $\omega$. Horizontal-axis: values of $\omega$ ; Vertical-axis: the $R2$ on the validation set of AdaBoost regressor model.}\label{f1}
\end{figure}

\begin{center}
\begin{table}[htbp]\normalsize
 \caption{Comparison between the PSO-BO, L-BFGS-B-BO, TNC-BO } \centering

 \begin{tabular}{ccccccc}  \hline
Method & PSO-BO $(\omega=0.8)$ & L-BFGS-B-BO & TNC-BO & \\
\hline
MAX & \textbf {0.614} & 0.61 & 0.614 & \\
MIN & \textbf {0.603} & 0.595  & 0.598 & \\
AVE & \textbf {0.609} & 0.6021 & 0.6045 & \\ \hline
 \end{tabular}\label{}
\end{table}
\end{center}

\subsection{Hyper-parameter Optimization of RF}

In this section, we estimated four hyper-parameters $(d=4)$ of RF classifier using Digits dataset. And the four hyper-parameters were estimated by PSO-BO, L-BFGS-B-BO and TNC-BO respectively.  The range and type of hyper-parameters be optimized is displayed in Table $1$. Note that while the first parameter takes real values, the others take integer values. As shown in Table $4$, comparing the averaged results, maximum results and minimum results, PSO-BO outperforms the other two algorithms under comparison in terms of the accuracy of classification on the validation set of the Digits dataset, and $5$-fold cross validation was used on the dataset. In Fig.$3$, the vertical axis represents the accuracy,  it can be seen that PSO-BO with $\omega=0.8$  performed better than the other settings.

\begin{figure}[!htb]
\centering
  \includegraphics[width=245pt]{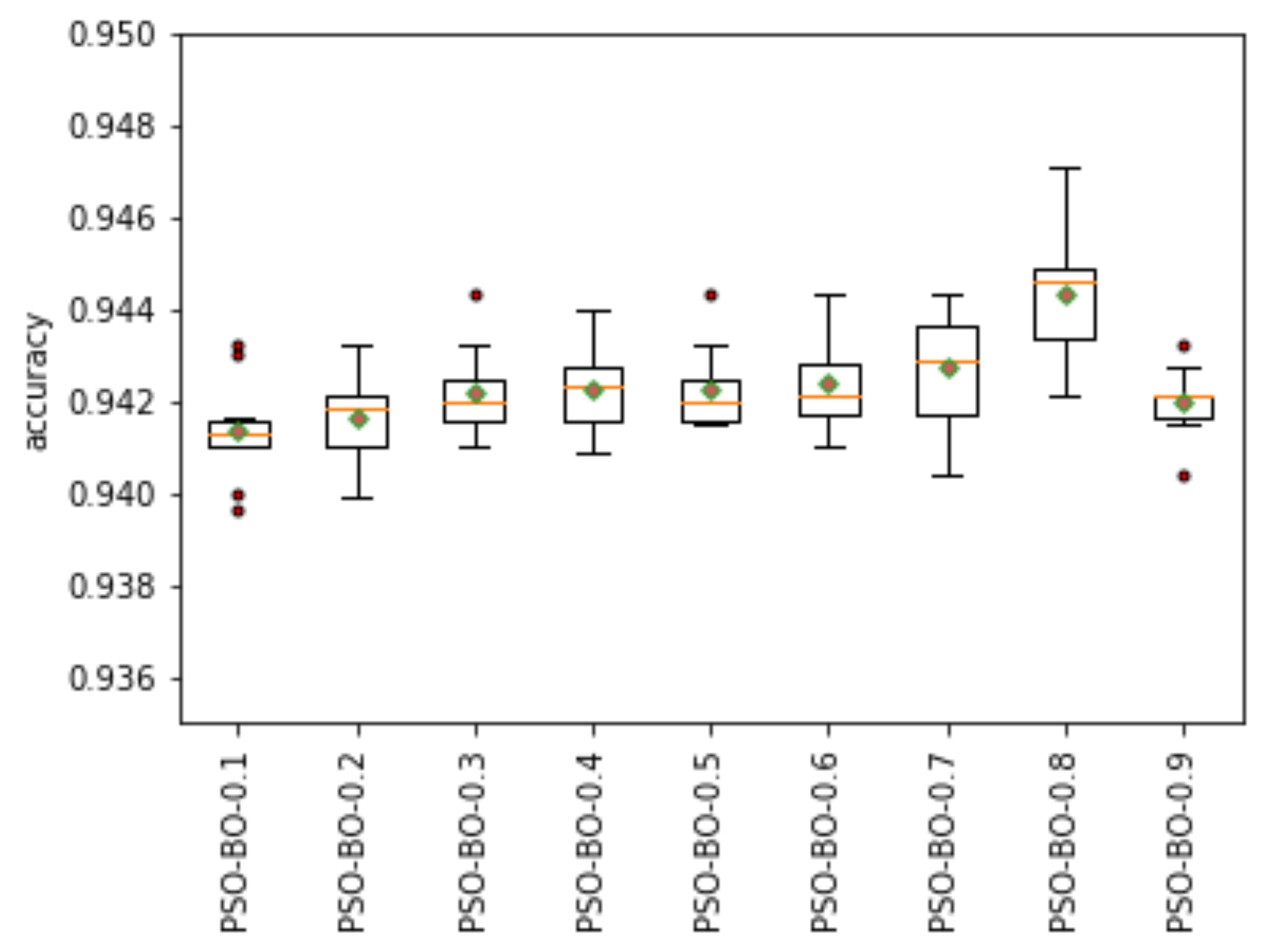}\\
  \caption{Experimental results of the developed PSO-BO approach with different values of $\omega$.  Horizontal-axis: values of $\omega$ ; Vertical-axis: the $accuracy$ on the validation set of RF classifier model.}\label{f1}
\end{figure}

\begin{center}
\begin{table}[htbp]\normalsize
 \caption{Comparison between the PSO-BO, L-BFGS-B-BO, TNC-BO } \centering

 \begin{tabular}{ccccccc}  \hline
Method & PSO-BO $(\omega=0.8)$ & L-BFGS-B-BO & TNC-BO & \\
\hline
MAX & \textbf{0.9471} & 0.9449 & 0.9443 & \\
MIN & \textbf{0.9421} & 0.9404 & 0.9399 & \\
AVE & \textbf{0.9443} & 0.9427 & 0.9421 & \\ \hline
 \end{tabular}\label{}
\end{table}
\end{center}

\subsection{Hyper-parameter Optimization of XGBoost}

In this section, we estimated five hyper-parameters $(d=5)$ of XGBoost classifier using Digits dataset. And the five hyper-parameters are estimated by PSO-BO, L-BFGS-B-BO and TNC-BO respectively. The range and type of hyper-parameters be optimized is displayed in Table $5$. Note that while the first and second parameters take real values, the others take integer values. As shown in Table $6$, comparing the averaged results, maximum results and minimum results, PSO-BO outperforms the other two algorithms under comparison in terms of the accuracy of classification on the validation set of the Digits dataset, and $5$-fold cross validation was used on the dataset. In Fig.$4$, the vertical axis represents the accuracy,  it can be seen that PSO-BO with $\omega=0.8$  performed better than the other settings.

\begin{center}
\begin{table}[htbp]\normalsize
 \caption{Types and range of hyper-parameters of XGBoost } \centering

 \begin{tabular}{ccccccc}  \hline
\textbf{Name} & \textbf{Type}  & \textbf{Range} & \\
\hline
Sub sample & Real  & (0.5, 1) & \\
Col sample by tree & Real  & (0.1, 1) & \\
Gamma & Real  & (0, 10) & \\
Min child weight & Integer & (1, 20) & \\
Max depth & Integer & (2, 10) & \\ \hline
 \end{tabular}\label{}
\end{table}
\end{center}

\begin{figure}[!htb]
\centering
  \includegraphics[width=245pt]{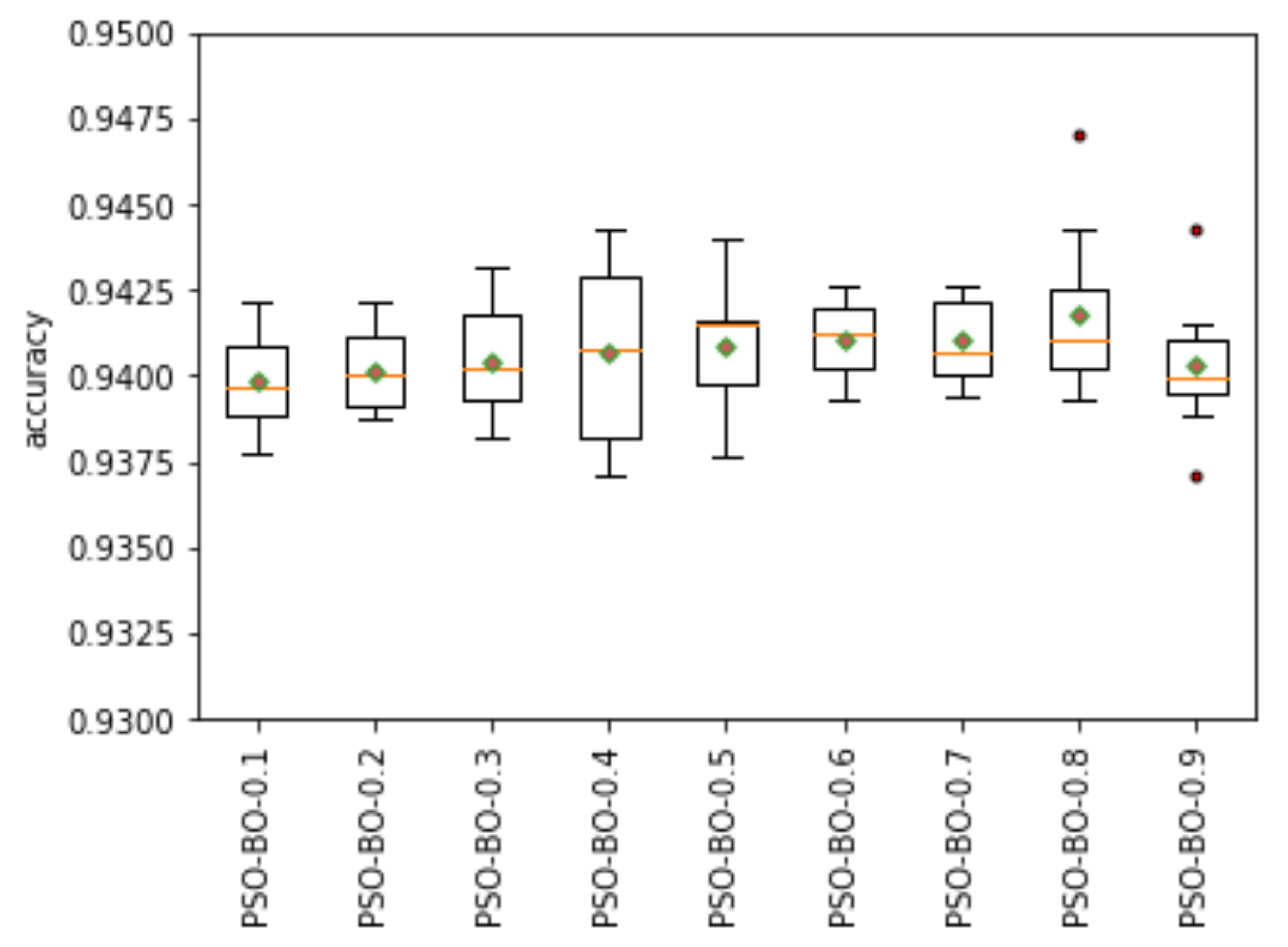}\\
  \caption{Experimental results of the developed PSO-BO approach with different values of $\omega$.  Horizontal-axis: values of $\omega$ ;  Vertical-axis: the $accuracy$ on the validation set of XGBoost classifier model.}\label{f1}
\end{figure}

\begin{center}
\begin{table}[htbp]\normalsize
 \caption{Comparison between the PSO-BO, L-BFGS-B-BO, TNC-BO } \centering

 \begin{tabular}{ccccccc}  \hline
Method & PSO-BO $(\omega=0.8)$ & L-BFGS-B-BO & TNC-BO & \\
\hline
MAX & \textbf{0.947} & 0.9438 & 0.946 & \\
MIN & \textbf{0.9393} & 0.9376 & 0.938 & \\
AVE & \textbf{0.9414} & 0.93945 & 0.9405 & \\ \hline
 \end{tabular}\label{}
\end{table}
\end{center}

\section{Conclusion}
In this paper, we developed a new approach, PSO-BO, based on PSO algorithm. In PSO-BO framework, PSO method is used to optimize the acquisition function to obtain new evaluation points, that significantly reduces the computational burden. Empirical evaluation on machine learning model showed that PSO-BO improves upon the state of the art. The resulting method can be used with most acquisition function. However, the algorithm runs slowly in high-dimensional space. In the future work, we plan to improve the PSO-BO algorithm on this basis to improve its running efficiency in high-dimensional space.

\section*{Acknowledgements}
This work is supported by NSFC-61803337, NSFC-61803338, ZJSTF-LGF18F020011.

\bibliographystyle{elsarticle-num}
\bibliography{eaai1611}

\end{document}